\documentclass[runningheads]{llncs}

\usepackage[T1]{fontenc}
\usepackage{graphicx}
\usepackage{tikz}
\usetikzlibrary{arrows.meta, positioning, fit, backgrounds, calc}

\usepackage{amsmath}
\usepackage{amssymb}

\usepackage{xurl}
\usepackage{hyperref}

\usepackage{placeins}

\definecolor{nodeblue}{RGB}{24,95,165}
\definecolor{nodebluefill}{RGB}{230,241,251}
\definecolor{nodepurple}{RGB}{83,74,183}
\definecolor{nodepurplefill}{RGB}{238,237,254}
\definecolor{nodeteal}{RGB}{15,110,86}
\definecolor{nodetealf}{RGB}{225,245,238}
\definecolor{nodecoral}{RGB}{153,60,29}
\definecolor{nodecoralfill}{RGB}{250,236,231}
\definecolor{annotgray}{RGB}{105,105,105}
\definecolor{groupfill}{RGB}{245,245,245}

\begin{document}

\title{EchoRisk: A Multicentre Echocardiography Dataset and Benchmark for Cardio-Oncology}

\titlerunning{EchoRisk: A Benchmark for Cardio-Oncology}

\author{
Grigorios Kalliatakis\inst{1}\thanks{Corresponding author: gkalliatak@ics.forth.gr} \and
Georgia Karanasiou\inst{2} \and
Georgios Manikis\inst{1} \and
Manolis Tsiknakis \inst{3} \and
Dimitrios Fotiadis \inst{2} \and
Dorothea Tsekoura \inst{4} \and
Kalliopi Keramida \inst{4} \and
Vasileios Bouratzis \inst{2} \and
Lampros Lakkas \inst{2} \and
Katerina Naka \inst{2} \and
Andri Papakonstantinou \inst{5} \and
Anastasia Constantinidou \inst{6} \and
Kostas Marias\inst{1}
}

\authorrunning{Grigorios Kalliatakis et al.}

\institute{
Foundation for Research and Technology Hellas, Greece\\
\and
University of Ioannina, Greece
\and
Hellenic Mediterranean University, Greece
\and
National and Kapodistrian University of Athens, Greece
\and
Karolinska University Hospital, Sweden
\and
Bank of Cyprus Oncology Centre, Cyprus
}

\maketitle

\begin{abstract}
Therapy-induced cardiotoxicity is the leading non-oncological cause of treatment interruption in breast cancer patients, yet early, automated risk stratification from routine cardiac imaging remains an unsolved problem. We present EchoRisk, the first curated, multicentre, longitudinal echocardiography dataset with explicit cardiotoxicity labels, released as the primary technical reference for the EchoRisk-MICCAI 2026 challenge. The dataset comprises 422 patients enrolled in the EU-funded CARDIOCARE prospective study across five European sites, yielding 2,159 echocardiography videos across 1,123 clinical exams acquired at up to five longitudinal timepoints, alongside a dedicated cohort of 280 patients with baseline imaging for early cardiotoxicity prediction. Three clinically grounded tasks are defined: automated estimation of left ventricular ejection fraction from cine video (Task~1), classification of LV dysfunction from longitudinal imaging (Task~2), and early prediction of therapy-induced cardiotoxicity from pre-therapy baseline echocardiography alone (Task~3). For each task we specify the evaluation protocol, primary and secondary metrics, and ranking procedure. We establish baseline performance using an R(2+1)D video backbone with LSTM aggregation trained from Kinetics-400 pretrained weights, demonstrating strong discriminative performance for cardiac functional assessment and LV dysfunction classification, while early cardiotoxicity prediction from a single pre-therapy video remains a significant open problem for the community. The dataset, evaluation code, and baseline implementations are publicly available to serve as a benchmark for further collaboration, comparison, and the creation of task-specific architectures in cardio-oncology.

\end{abstract}


\section{Introduction}
\label{sec:introduction}

Breast cancer treatment has advanced substantially over the past two decades, yet the cardiotoxic side effects of established therapies remain a critical and growing clinical concern. Anthracyclines and HER2-targeted agents, such as trastuzumab, induce cardiac dysfunction in up to 20--30\% and 7--10\% of patients, respectively~\cite{zamorano2016esc,lyon2020hfaicos}. Cardiotoxicity is not merely an incidental complication: it is the leading non-oncological cause of treatment interruption and long-term morbidity in this population, with subclinical dysfunction linked to elevated cardiovascular mortality even years after therapy completion~\cite{cardinale2015early}. Early detection and risk stratification are therefore essential, enabling timely cardioprotective intervention and personalised surveillance before irreversible damage occurs.

Echocardiography is the standard of care for cardiac monitoring in cancer patients, providing non-invasive, radiation-free assessment of left ventricular structure and function at multiple longitudinal timepoints. Despite its central role, clinical decision-making based on echocardiography faces two fundamental limitations. First, manual analysis is operator-dependent and subject to significant intra- and inter-observer variability, with ejection fraction measurements varying by up to 5--10 percentage points between readers~\cite{thavendiranathan2013reproducibility}. Second, conventional surveillance relies on detecting a meaningful drop in ejection fraction, a relatively late indicator of cardiac injury. Deep learning methods applied directly to echocardiography video have the potential to address both limitations by providing automated, reproducible estimates of cardiac function and, crucially, by extracting predictive imaging features not captured by current clinical indices.

Progress in this direction has been demonstrated by EchoNet-Dynamic~\cite{ouyang2020video}, which established that spatiotemporal convolutional architectures can estimate ejection fraction from apical four-chamber video at near-expert accuracy on a dataset of over 10,000 patients from a single centre. Subsequent work has extended video-based analysis to view classification, segmentation, and volume estimation~\cite{leclerc2019deep,ghorbani2020deep}. However, no existing public echocardiography dataset or benchmark addresses the specific clinical problem of therapy-induced cardiotoxicity in a longitudinal, multicentre setting. Existing resources are limited to single-site cohorts, generic cardiac function tasks, and absence of cardiotoxicity-specific labels linked to long-term follow-up.

\begin{figure}[!t]
    \centering
    \resizebox{\textwidth}{!}{%
    \begin{tikzpicture}[
        base/.style={draw, thick, align=center, inner sep=1.5ex, rounded corners=2pt},
        echo/.style={base, fill=blue!5, draw=blue!80!black, minimum height=1cm},
        task1/.style={base, fill=teal!5, draw=teal!80!black, minimum width=3.8cm},
        task2/.style={base, fill=orange!5, draw=orange!80!black},
        task3/.style={base, fill=red!5, draw=red!80!black},
        arrow/.style={-{Stealth[scale=1.2]}, thick},
        dashed_arrow/.style={-{Stealth[scale=1.2]}, thick, dashed},
        timeline/.style={thick, -{Stealth[scale=1.2]}}
    ]

    \draw[timeline] (0,0) -- (12.5,0) node[right] {\textbf{Time}};

    \foreach \x/\tlabel in {1.0/{$T_1$ \\ (Baseline)}, 3.5/{$T_2$ \\ (Month 3)}, 6.0/{$T_3$ \\ (Month 6)}, 8.5/{$T_4$ \\ (Month 9)}, 11.0/{$T_5$ \\ (Month 12)}} {
        \draw[thick] (\x,0.15) -- (\x,-0.15) node[below=0.1cm, align=center] {\footnotesize \tlabel};
    }

    \node[echo, minimum width=1.8cm] (e0) at (1.0, 1.4) {Echo \\ \footnotesize Pre-therapy};
    \node[echo, minimum width=8.2cm] (eduring) at (7.25, 1.4) {Echo \\ \footnotesize During Therapy};

    \begin{scope}[on background layer]
        \node[draw=gray!80, thick, dashed, fill=gray!10, inner xsep=0.3cm, inner ysep=0.4cm, fit=(e0) (eduring)] (longbox) {};
    \end{scope}

    \node[task2] (t2) at (9.0, 3.2) {\textbf{Task 2} \\ Assess \\ \footnotesize (LV dysfunction)};
    \draw[arrow, draw=orange!80!black] (eduring.north -| 9.0,0) -- (9.0,0 |- t2.south);

    \node[task1] (t1) at (5.0, 3.2) {\textbf{Task 1} \\ Measure \\ \footnotesize (LVEF estimation)};
    \draw[arrow, draw=teal!80!black] (eduring.north -| 5.0,0) -- (5.0,0 |- t1.south);

    \node[task3] (t3) at (1.0, 4.8) {\textbf{Task 3} \\ Forecast \\ \footnotesize (Cardiotoxicity Prediction)};
    \draw[arrow, draw=red!80!black] (e0.north) -- (t3.south);

    \draw[dashed_arrow, draw=red!80!black] (t3.east) -- node[above] {\footnotesize Predicts future cardiotoxicity development} (11.5, 4.8);

    \end{tikzpicture}%
    }
    \caption{Clinical formulation of the EchoRisk-MICCAI 2026 challenge tasks. Task 1 focuses on cross-sectional estimation of cardiac function at any given timepoint. Task 2 tracks these parameters across multiple timepoints to classify longitudinal LV dysfunction. Task 3 utilises only the baseline (pre-therapy) echocardiogram to predict the future risk of developing cardiotoxicity.}
    \label{fig:clinical_tasks}
\end{figure}
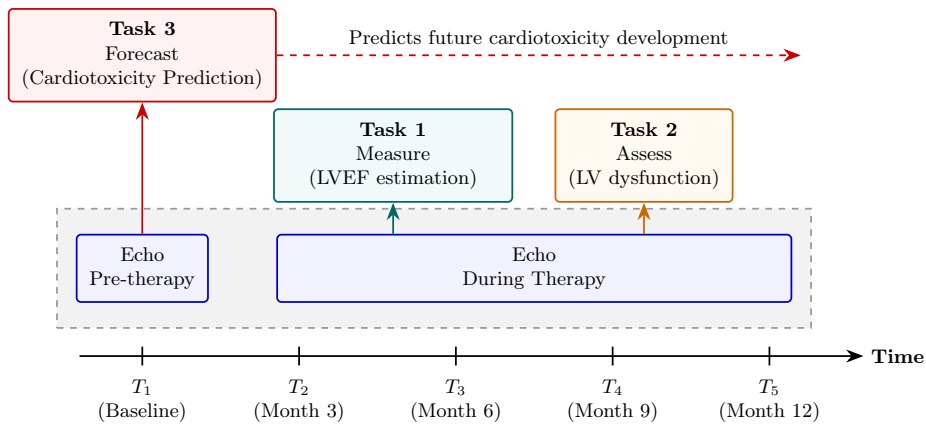

EchoRisk is built on the EU-funded CARDIOCARE project~\cite{cardiocare}, a prospective clinical study enrolling breast cancer patients undergoing cardiotoxic therapy across five European sites in four countries, which has already demonstrated the prognostic value of echocardiographic radiomics for this specific patient population~\cite{manikis2025association}. Three clinically grounded tasks are defined: automated estimation of cardiac functional parameters (Task~1), LV dysfunction assessment from longitudinal imaging (Task~2), and early prediction of therapy-induced cardiotoxicity from baseline echocardiography alone (Task~3; see Fig.~\ref{fig:clinical_tasks}). Full dataset statistics, task definitions, and evaluation protocols are described in Sections~2 and~3; baseline results are reported in Section~4. 

This paper serves as the primary technical reference for the EchoRisk-MICCAI 2026 challenge. Our contributions are as follows. We describe the EchoRisk dataset, its annotation protocol, and the evaluation infrastructure underpinning the challenge. We establish baseline performance across all three tasks using an R2+1D video backbone with LSTM aggregation, trained from Kinetics-400 pretrained weights and evaluated across eight independent random seeds (seeds 42--49). For Task~3, we additionally establish a clinical reference floor using logistic regression on age and baseline LVEF, aligned with the HFA-ICOS risk stratification framework~\cite{lyon2020hfaicos}. Together, these baselines map the current performance bounds for video-based approaches on each task and provide a reference against which challenge participants can measure their contributions.

\section{The EchoRisk Dataset}
\label{sec:dataset}

\subsection{Source cohort}
\label{subsec:source_cohort}

EchoRisk is built on the CARDIOCARE prospective multicentre clinical study~\cite{cardiocare}, an EU Horizon 2020 project designed to develop an integrated, interdisciplinary approach to the management of elderly multimorbid patients with breast cancer therapy-induced cardiac toxicity. Patients with early-stage breast cancer scheduled to receive cardiotoxic treatment were enrolled across five European sites in four countries. Ethical approval was obtained at each participating site in accordance with national and local regulations. All data are fully de-identified and GDPR-compliant prior to release. All dataset statistics reported in this document correspond to EchoRisk data release v1.0 (April 2026).

\subsection{Inclusion and exclusion criteria} 
\label{subsec:inclusion_exclusion} 
The imaging release includes all patients from the CARDIOCARE prospective cohort for whom at least one echocardiography acquisition was available at baseline ($T_1$) with an associated clinical label. No cases were excluded on the basis of image quality: acquisitions classified as good, medium, or poor by the acquiring sonographer are retained in the release, reflecting the full range of real-world clinical image quality encountered in routine cardiotoxicity surveillance. Cases were excluded only if (i)~no DICOM file was recoverable for any view at any timepoint, or (ii)~the patient withdrew consent prior to data transfer. Scanner text overlays, acquisition artefacts, and suboptimal probe orientations are retained without correction.

\subsection{Imaging protocol}
\label{subsec:imaging_protocol}

\begin{figure}[t]
\centering
\includegraphics[width=\textwidth]{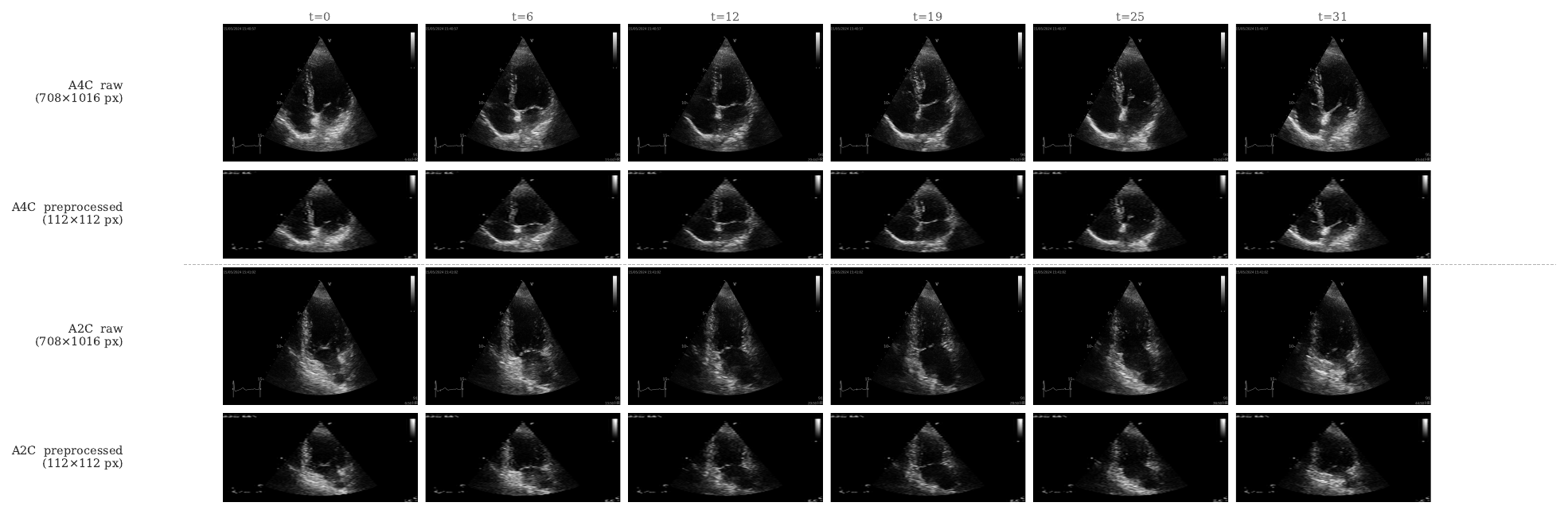}
\caption{Representative echocardiography frames from a single patient (ECHORISK\_0001, baseline $T_1$ visit). Each row shows six of 32 evenly sampled frames spanning one full cardiac cycle. \textit{Top pair:} apical four-chamber (A4C) view. \textit{Bottom pair:} apical two-chamber (A2C) view. Within each pair, raw DICOM frames (native resolution, upper row) are shown alongside preprocessed frames (lower row) after resizing to $112\times112$ pixels and per-clip z-score normalisation. Note the aspect ratio change introduced by resizing non-square native frames to the fixed spatial resolution required by the Kinetics-400 pretrained backbone. Scanner text overlays and acquisition artefacts are retained in the released DICOMs.}
\label{fig:echo_panel}
\end{figure}

Echocardiography examinations were conducted as part of routine clinical care following standard acquisition protocols at each participating centre. For each patient, two-dimensional (2D) greyscale cine echocardiography was acquired in the apical four-chamber (A4C) and apical two-chamber (A2C) views using multi-vendor ultrasound systems (GE, Philips, Siemens, Canon). No challenge-specific imaging protocol was imposed; acquisitions reflect real-world clinical conditions including variability in frame rate, spatial resolution, and probe orientation. Images are provided in their original DICOM format without any preprocessing, preserving all native acquisition characteristics and metadata. At least one full cardiac cycle was acquired per view per examination. Representative examples of native and preprocessed frames are shown in Fig.~\ref{fig:echo_panel}.

For the longitudinal tasks (Tasks~1 and~2), up to five timepoints were collected per patient at approximately three-month intervals corresponding to the baseline visit ($T_1$) and follow-up visits at 3, 6, 9, and 12~months ($T_2$--$T_5$). For the early cardiotoxicity prediction task (Task~3), only the baseline $T_1$ examination is used as model input, with the cardiotoxicity outcome determined from longitudinal follow-up.

\begin{table}[!b]
\centering
\caption{Native DICOM frame count per clip for Tasks~1 and~2
(2,159 clips from 1,123 exams; 1,036 exams (92\%) have both views,
72 A4C-only and 15 A2C-only). All timepoints combined,
prior to resampling. IQR: interquartile range.}
\label{tab:framecounts}
\setlength{\tabcolsep}{8pt}
\renewcommand{\arraystretch}{1.3}
\begin{tabular}{lrrrr}
\hline
\textbf{View} & \textbf{Clips} & \textbf{Median (IQR)} & \textbf{Min} & \textbf{Max} \\
\hline
A4C             & 1{,}108 & 60 (44--154) & 15  & 711 \\
A2C             & 1{,}051 & 61 (44--158) & 14  & 574 \\
\hline
Both            & 2{,}159 & 60 (44--156) & 14  & 711 \\
\hline
\end{tabular}
\end{table}

Native clip lengths vary substantially across vendors and acquisition sites. The median frame count is 60 (IQR 44--156, range 14--711 across both views), with 18.4\% of clips exceeding 200~frames owing to extended multi-cycle acquisitions at higher-frame-rate sites, and 5.4\% containing fewer than 30~frames. The maximum observed clip length (711~frames, \texttt{ECHORISK\_0407\_T2\_A4C}) corresponds to an extended acquisition spanning approximately 8--10 cardiac cycles, consistent with one participating site's clinical acquisition protocol. Summary statistics are reported in Table~\ref{tab:framecounts}. Fractional index sampling (Section~\ref{subsec:model_architectures}) normalises all clips to $T\!=\!32$ frames prior to model input, regardless of native frame count.

\subsection{Annotation protocol}
\label{subsec:annotation_protocol}

\subsubsection{Tasks 1 and 2: cardiac parameter estimation and LV dysfunction.}
Left ventricular ejection fraction (LVEF) and global longitudinal strain (GLS) were measured as part of standard clinical echocardiography workflows at each participating centre using vendor-provided or clinically validated software, in accordance with the guidelines of the American Society of Echocardiography and the European Association of Cardiovascular Imaging~\cite{lang2015recommendations}. LVEF was derived by the biplane Simpson's method from A4C and A2C views. GLS was used as the primary label source for Task~2, with LVEF serving as a fallback for examinations in which reliable GLS measurement was not available. LV dysfunction is defined as GLS $\geq -16\%$ primary or LVEF $< 50\%$ fallback, consistent with the ESC 2022 guidelines on cardio-oncology~\cite{lyon2022esc}. Of the 1123 examinations in the Tasks~1 and~2 cohort, approximately 85\% carry GLS-derived dysfunction labels and 15\% carry LVEF-derived fallback labels owing to unreliable GLS acquisition at specific timepoints or sites. This proportion is consistent across training, validation, and test splits. All measurements reflect real-world inter-reader variability across participating centres; no centralised re-annotation was performed for the challenge.

\subsubsection{Task 3: early prediction of cardiotoxicity.}
Cardiotoxicity outcome labels were established through a structured expert adjudication process conducted by domain experts at each participating centre, integrating longitudinal echocardiography findings with cardiotoxicity-related blood biomarkers including troponin and NT-proBNP. Adjudication followed established clinical criteria based on the ESC 2022 cardio-oncology guidelines~\cite{lyon2022esc}. This expert-curated labelling framework represents a deliberate, labour-intensive curation process rather than routine clinical reporting or automated extraction from electronic health records. Labels reflect the presence or absence of therapy-induced cardiotoxicity within each patient's observed follow-up window at the time of the current data cut. The CARDIOCARE study is ongoing; for patients enrolled more recently, follow-up extending to 12~months had not yet been reached at the time of dataset release. Outcome labels therefore correspond to the latest available adjudicated timepoint per patient, which ranges from $T_2$ (3~months) to $T_5$ (12~months). No censoring imputation is applied. Participants should interpret Task~3 AUC estimates within this context: the positive label indicates cardiotoxicity confirmed within the available window, not necessarily within a fixed 12-month horizon. This constraint reflects inherent limitations of prospective longitudinal data collection and mirrors real-world clinical practice. Importantly, all reference labels across training, validation, and test sets were derived using the same standardised adjudication process, ensuring internal consistency. Participants are not required to model the outcome horizon explicitly; the task is defined as binary classification at the patient level using the available adjudicated label, consistent with the evaluation protocol in Section~3.3.

\subsection{Dataset statistics}
\label{subsec:dataset_statistics}

The EchoRisk dataset encompasses two overlapping imaging cohorts. The first supports Tasks~1 and~2 and comprises 422 patients with 2,159 echocardiography videos across 1,123 multi-timepoint examinations. The second supports Task~3 and comprises 280 patients, each with a single baseline examination (560 videos). Both views (A4C and A2C) are available for 92\% of Task~1/2 exams and 100\% of Task~3 exams. Table~\ref{tab:dataset} summarises the key dataset characteristics across all three tasks. The dataset is distributed across the participating European sites, reflecting real-world variance in scanner hardware and clinical populations. Class distribution in Task~2 reflects the real-world prevalence of LV dysfunction in this population (16.8\% positive rate). Task~3 carries a majority-positive class (65.4\%), consistent with the enriched high-risk enrolment criteria of the CARDIOCARE study, in which patients received known cardiotoxic regimens.

\begin{table}[t]
\centering
\caption{EchoRisk dataset overview. All splits are at the patient level. Task~3 exams correspond to the T1 baseline timepoint only.}
\label{tab:dataset}
\renewcommand{\arraystretch}{1.2}
\begin{tabular}{lccc}
\hline
& \textbf{Task 1} & \textbf{Task 2} & \textbf{Task 3} \\
& LVEF Estimation & LV Dysfunction & Cardiotoxicity \\
\hline
Patients & 422 & 422 & 280 \\
Videos & 2,159 & 2,159 & 560 \\
Exams & 1,123 & 1,123 & 280 \\
Timepoints & T1--T5 & T1--T5 & T1 only \\
Both views (\%) & 92 & 92 & 100 \\
\hline
Train (patients / exams) & 237 / 630 & 237 / 630 & 157 / 157 \\
Val (patients / exams) & 59 / 156 & 59 / 156 & 39 / 39 \\
Test (patients / exams) & 126 / 337 & 126 / 337 & 84 / 84 \\
\hline
Positive rate & --- & 16.8\% & 65.4\% \\
Primary metric & MAE (pp) & AUC-ROC & AUC-ROC \\
\hline
\end{tabular}
\end{table}

\section{Tasks and Evaluation Protocol}
\label{sec:tasks_and_evaluation_protocol}

EchoRisk introduces three clinically grounded tasks of increasing difficulty. The evaluation framework is explicitly modular; each task is assessed independently to isolate algorithmic contributions to specific clinical problems. Performance metrics are calculated separately for each task, ensuring that specialized architectures can be robustly evaluated within their designated clinical context without the need for cross-task aggregation.

\subsection{Task 1: Cardiac Parameter Estimation}
\label{subsec:task_1_cardiac_parameter_estimation}

\paragraph{Problem formulation.}
Given a 2D echocardiography cine video from any available timepoint ($T_1$--$T_5$), automatically estimate left ventricular ejection fraction (LVEF, \%). Models are trained and evaluated on the Tasks~1 and~2 cohort (422 patients, 1,123 examinations; Table~\ref{tab:dataset}). Algorithms must output a single continuous scalar value per examination. The required submission granularity is one prediction per examination; participants are free to fuse multiple views prior to producing a single scalar output.

\paragraph{Primary metric and ranking.}
Mean absolute error (MAE) is the primary metric, providing a clinically interpretable measure of error in native percentage points (pp). An absolute LVEF error within 5~pp is considered clinically acceptable; deviations beyond 10~pp may alter management around key decision thresholds~\cite{lang2015recommendations}. Algorithms are ranked in ascending order of MAE. Ties are resolved by lower Root Mean Squared Error (RMSE).

\paragraph{Secondary metrics.}
Root mean squared error (RMSE), Pearson correlation ($r$), and the coefficient of determination ($R^2$) are reported as complementary metrics to assess outlier sensitivity and overall correlation.

\subsection{Task 2: LV Dysfunction Classification}
\label{subsec:task_2_lv_dysfunction_classification}

\paragraph{Problem formulation.}
Given a 2D echocardiography cine video from any available timepoint ($T_1$--$T_5$), output a probability score of LV dysfunction in $[0,1]$ per examination. LV dysfunction is defined as GLS $\geq -16\%$ (primary criterion) or LVEF $< 50\%$ (fallback criterion), consistent with the ESC 2022 cardio-oncology guidelines~\cite{lyon2022esc}. The overall positive rate is 16.8\%, reflecting real-world clinical prevalence in a cardiotoxicity surveillance population. The required submission granularity is one probability score per examination.

\paragraph{Clinical rationale.}
GLS is a more sensitive and earlier marker of subclinical cardiac dysfunction than LVEF, detecting myocardial deformation abnormalities before ejection fraction decline becomes apparent~\cite{zamorano2016esc}. Automating GLS-defined dysfunction detection from raw cine video bypasses manual strain analysis and reduces inter-observer variability~\cite{thavendiranathan2013reproducibility}.

\paragraph{Primary metric and ranking.}
Area under the receiver operating characteristic curve (AUC-ROC) is the primary metric. It is threshold-independent and measures discriminative ability across all operating points. Algorithms are ranked in descending order of AUC. Ties are resolved by (1)~higher sensitivity at 90\% specificity, then (2)~higher balanced accuracy.

\subsection{Task 3: Early Cardiotoxicity Prediction}
\label{subsec:task_3_early_cardiotoxicity_prediction}

\paragraph{Problem formulation.}
Given only the baseline ($T_1$, pre-therapy) echocardiography video, output a probability score $p \in [0,1]$ of developing therapy-induced cardiotoxicity within the available follow-up window (up to 12~months). Task~3 uses a dedicated cohort of 280 patients with baseline-only imaging (Table~\ref{tab:dataset}). No follow-up echocardiography or clinical variables are provided at inference time. The overall positive rate is 65.4\%, reflecting enriched high-risk enrolment in the CARDIOCARE study. The required submission granularity is one probability score per patient, as each patient contributes exactly one baseline examination in this cohort.

\paragraph{Predictive complexity.}
Tasks~1 and~2 ask models to characterise the cardiac state visible in the input video; the signal of interest is present at acquisition time. Task~3 asks models to predict a future outcome from a single baseline video acquired before any cardiac injury has occurred. The predictive signal is subtle, subclinical, and may not be perceptible from spatiotemporal video representations alone at the training scales considered here, making this an open problem for the community; handcrafted radiomics features extracted from the same cohort have shown statistically significant association with cardiotoxicity outcome~\cite{manikis2025association}, suggesting that the relevant prognostic signal is present in the imaging data even where end-to-end video architectures at the current training scale fail to recover it.

\paragraph{Primary metric and ranking.}
AUC-ROC is the primary metric. Algorithms are ranked in descending order. Ties are resolved sequentially by (1)~higher sensitivity at a false positive rate (FPR) $\leq 0.20$, (2)~higher balanced accuracy, and (3)~lower Brier score~\cite{steyerberg2019clinical}. The Brier score is defined as $\mathrm{BS} = \frac{1}{N}\sum_{i=1}^{N}(p_i - y_i)^2$, where $p_i \in [0,1]$ is the predicted probability and $y_i \in \{0,1\}$ is the binary outcome label. Lower values indicate better calibration and probabilistic accuracy.

\paragraph{Clinical target.}
The clinical target for early risk screening is sensitivity $\geq 0.80$ at an FPR of 0.10--0.20, enabling timely cardioprotective intervention while maintaining acceptable specificity~\cite{lyon2020hfaicos}.

\paragraph{Clinical reference baseline.}
A clinical reference derived from age and baseline LVEF (logistic regression, aligned with the HFA-ICOS risk stratification framework~\cite{lyon2020hfaicos}) is evaluated internally using secure CARDIOCARE registry variables. This baseline is excluded from algorithmic rankings and is included solely to contextualise imaging-based predictive performance relative to guideline-aligned clinical practice.

\subsection{Statistical Analysis and Reproducibility}
\label{subsec:statistical_analysis_and_reproducibility}

To ensure standardised and reproducible performance assessment, the
official evaluation protocol and scoring library are publicly available
at \url{https://github.com/EchoRisk-MICCAI/echorisk-benchmark}. Models
failing to produce a valid numerical output for one or more test cases
are penalised as follows. For Task~1 (regression), missing or non-finite
predictions are assigned an absolute error of 100 percentage points for
those cases prior to MAE aggregation. For Tasks~2 and~3
(classification), missing predictions are assigned a score of 0 for AUC
computation purposes, consistent with random-chance behaviour.

Performance estimates are reported with 95\% confidence intervals
derived from non-parametric percentile bootstrap resampling across the
test sets. To establish statistical significance when comparing distinct
algorithmic approaches, pairwise comparisons utilise the two-sided
Wilcoxon signed-rank test. The Holm-Bonferroni correction is applied to
control the family-wise error rate across multiple comparisons.

For Tasks~2 and~3, calibration is assessed via the Expected Calibration
Error (ECE), computed as:
\begin{equation}
  \mathrm{ECE} = \sum_{b=1}^{B}
    \frac{|S_b|}{N}
    \left| \mathrm{acc}(S_b) - \mathrm{conf}(S_b) \right|,
\end{equation}
where predictions are partitioned into $B=10$ equal-width bins,
$|S_b|$ is the number of samples in bin $b$, $\mathrm{conf}(S_b)$ is
the mean predicted probability within the bin, and $\mathrm{acc}(S_b)$
is the fraction of positive cases within the
bin~\cite{naeini2015obtaining}. ECE is reported as a supplementary
diagnostic metric alongside a reliability diagram and does not affect
the official leaderboard ranking.

\section{Benchmark Performance}
\label{sec:benchmark_performance}

We establish reference implementations for all three tasks using a shared video backbone architecture. These reference models are designed to represent foundational deep learning approaches, mapping the current performance landscape across three principal design axes: clinical variables only, single-view video, and dual-view video.

\subsection{Model Architecture}
\label{subsec:model_architectures}

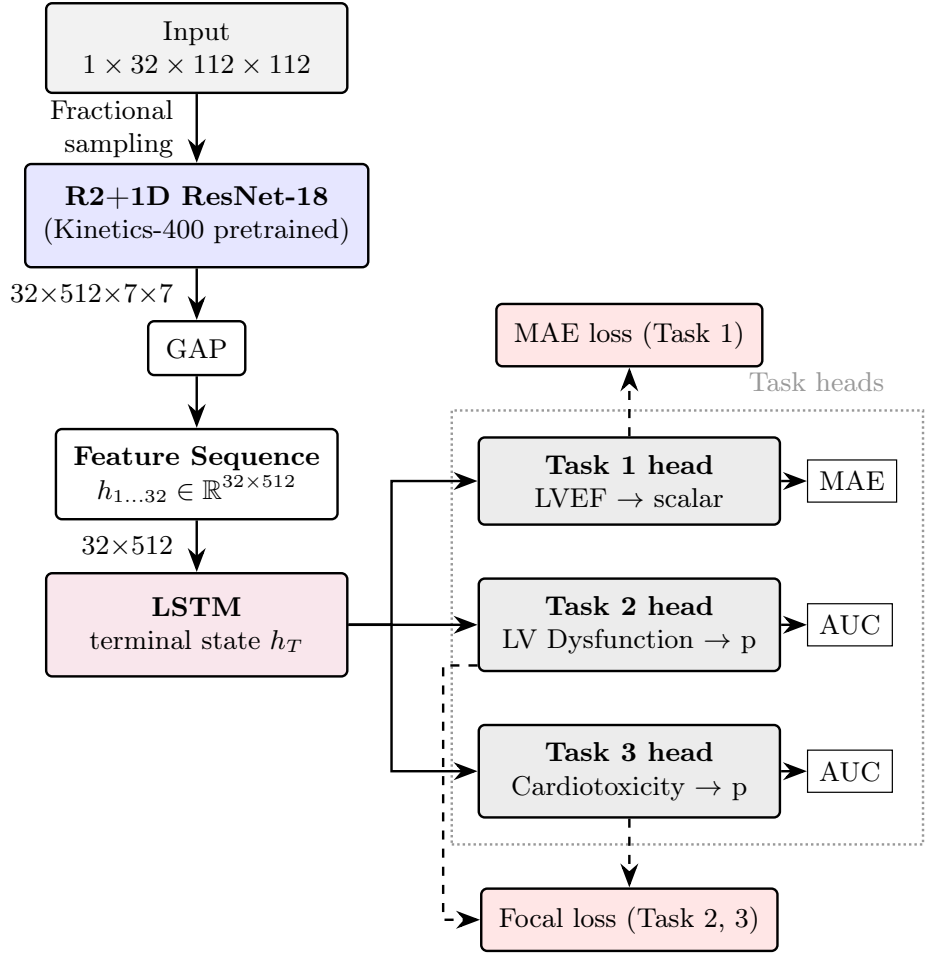
\begin{figure}[!ht]
    \centering
    \resizebox{\textwidth}{!}{%
    \begin{tikzpicture}[
        base/.style={draw, thick, align=center, inner sep=1.5ex, rounded corners=2pt},
        input/.style={base, fill=gray!10, minimum width=3.5cm},
        backbone/.style={base, fill=blue!10, minimum height=1.2cm, minimum width=3.5cm},
        aggregator/.style={base, fill=purple!10, minimum height=1.2cm, minimum width=3.5cm},
        task/.style={base, fill=gray!15, minimum width=3.5cm},
        loss/.style={base, fill=red!10, minimum width=3cm},
        metric/.style={draw, fill=white, inner sep=1ex},
        arrow/.style={-{Stealth[scale=1.2]}, thick},
        dashed_arrow/.style={-{Stealth[scale=1.2]}, thick, dashed}
    ]

    \node[input] (input) {Input \\ $1 \times 32 \times 112 \times 112$};
    \node[backbone, below=0.8cm of input] (backbone) {\textbf{R2+1D ResNet-18} \\ \footnotesize (Kinetics-400 pretrained)};
    \node[base, fill=white, below=0.6cm of backbone] (gap) {GAP};
    \node[base, fill=white, below=0.6cm of gap] (seq) {\textbf{Feature Sequence} \\ $h_{1...32} \in \mathbb{R}^{32 \times 512}$};
    \node[aggregator, below=0.6cm of seq] (lstm) {\textbf{LSTM} \\ \footnotesize terminal state $h_T$};

    \node[task, right=1.5cm of lstm] (task2) {\textbf{Task 2 head} \\ LV Dysfunction $\to$ p};
    \node[task, above=0.6cm of task2] (task1) {\textbf{Task 1 head} \\ LVEF $\to$ scalar};
    \node[task, below=0.6cm of task2] (task3) {\textbf{Task 3 head} \\ Cardiotoxicity $\to$ p};

    \node[metric, right=0.3cm of task1] (m1) {MAE};
    \node[metric, right=0.3cm of task2] (m2) {AUC};
    \node[metric, right=0.3cm of task3] (m3) {AUC};

    \node[loss, above=0.8cm of task1] (loss1) {MAE loss (Task 1)};
    \node[loss, below=0.8cm of task3] (loss23) {Focal loss (Task 2, 3)};

    \draw[arrow] (input) -- node[left=0.15cm, midway, align=right] {\footnotesize Fractional\\ \footnotesize sampling} (backbone);
    \draw[arrow] (backbone) -- node[left=0.15cm, midway] {\footnotesize $32{\times}512{\times}7{\times}7$} (gap);
    \draw[arrow] (gap) -- (seq);
    \draw[arrow] (seq) -- node[left=0.15cm, midway] {\footnotesize $32{\times}512$} (lstm);

    \draw[arrow] (lstm.east) -- ++(0.5,0) |- (task1.west);
    \draw[arrow] (lstm.east) -- (task2.west);
    \draw[arrow] (lstm.east) -- ++(0.5,0) |- (task3.west);

    \draw[arrow] (task1) -- (m1);
    \draw[arrow] (task2) -- (m2);
    \draw[arrow] (task3) -- (m3);

    \draw[dashed_arrow] (task1.north) -- (loss1.south);
    \draw[dashed_arrow] (task3.south) -- (loss23.north);
    \draw[dashed_arrow] (task2.195) -- ++(-0.4,0) |- (loss23.west);

    \begin{scope}[on background layer]
        \node[draw=gray!80, thick, densely dotted, inner sep=0.3cm, fit=(task1) (task3) (m1) (m3), label={[text=gray!80, xshift=1.5cm, yshift=0.1cm]above:Task heads}] {};
    \end{scope}
    \end{tikzpicture}%
    }
    \caption{EchoRisk baseline architecture. A shared R2+1D ResNet-18 backbone (Kinetics-400 pretrained~\cite{tran2018closer}) encodes a 32-frame echocardiography clip into a temporal feature sequence, which a single-layer LSTM aggregates into a fixed-length representation. Three task-specific linear heads branch from the LSTM terminal hidden state for LVEF regression (Task~1), LV dysfunction classification (Task~2), and cardiotoxicity prediction (Task~3). At inference, predictions from A4C and A2C views are averaged when both are available.}
    \label{fig:architecture}
\end{figure}

All video baselines share a common architecture illustrated in Fig.~\ref{fig:architecture}. A factorised spatiotemporal convolutional backbone (R2+1D ResNet-18, Kinetics-400 pretrained~\cite{tran2018closer}) encodes a $T=32$-frame greyscale clip of shape $(1, 32, 112, 112)$ into a temporal feature sequence. Spatial dimensions are collapsed by global average pooling (GAP)
prior to LSTM input, yielding a per-frame feature vector of
dimension 512. The backbone decomposes 3D convolution into a spatial 2D step followed by a temporal 1D step, improving parameter efficiency and optimisation stability relative to full 3D convolutions. A single-layer LSTM (hidden size 512) aggregates the temporal sequence into a fixed-length vector from the terminal hidden state, which a task-specific linear head then maps to the prediction: a scalar LVEF (Task~1), a sigmoid probability for LV dysfunction (Task~2), or a sigmoid probability for cardiotoxicity (Task~3).

The R2+1D backbone was pretrained on Kinetics-400 using RGB clips at $112\times112$ resolution. To accommodate single-channel greyscale echocardiography input, the three-channel convolutional stem is replaced by a single-channel stem initialised by averaging the pretrained RGB filter weights. Input frames are resized to $112\times112$ pixels to match the backbone's expected resolution, following the same preprocessing protocol as EchoNet-Dynamic~\cite{ouyang2020video}. This introduces mild aspect ratio distortion relative to native acquisition dimensions, which are typically non-square.

\paragraph{Frame sampling.}
Cardiac DICOM files do not carry reliable frame rate metadata in this dataset. We therefore apply fractional index sampling~\cite{ouyang2020video}: $T=32$ frame indices are drawn evenly from $[0, N-1]$ where $N$ is the total frame count reported in the DICOM header (\texttt{NumberOfFrames}), ensuring the sampled clip spans exactly one full cardiac cycle regardless of acquisition frame rate.

\paragraph{Multi-view strategy.}
Task~3 patients have both A4C and A2C views available (100\% dual-view coverage). For the single-view baseline, one view is selected randomly during training; at inference, predictions from both views are averaged. For the dual-view baseline, A4C and A2C clips are concatenated along the temporal axis before the backbone, producing a $(1, 64, 112, 112)$ input that gives the model simultaneous access to both geometric projections of the left ventricle.

\subsection{Implementation Details}
\label{subsec:implementation_details}

\subsubsection{Preprocessing.}
DICOM pixel arrays are decoded, averaged to greyscale when encoded as RGB, resampled to 32 frames via fractional index sampling, resized to $112\times112$ pixels by bilinear interpolation, and normalised by per-clip z-score. Preprocessed arrays are cached as \texttt{float32} NumPy files to eliminate DICOM decoding from the training bottleneck.

\subsubsection{Training.}
All models are trained with AdamW using a differential learning rate: $10^{-4}$ for the backbone and $10^{-3}$ for the LSTM and task head. Weight decay is $10^{-4}$ throughout. A linear warmup of 5 epochs is followed by cosine annealing over the remaining epochs, with early stopping at patience~15 on the primary validation metric. Automatic mixed precision (AMP) is used on all GPU runs.

\subsubsection{Losses.}
Task~1 uses mean absolute error. Task~2 uses weighted binary cross-entropy with
$w_\text{pos} = n_\text{neg}/n_\text{pos} = 4.575$, derived from the training split
class frequencies (LV dysfunction is under-represented relative to the overall
16.8\% positive rate, consistent with the imbalance observed in Task~3's training
split). Task~3 uses Focal Loss~\cite{lin2017focal} with $\alpha=0.325$ and
$\gamma=2.0$. The alpha value equals
$n_\text{neg} / (n_\text{pos} + n_\text{neg}) = 51/157 = 0.325$, down-weighting the
majority positive class (67.5\% of training cases).

\subsubsection{Reproducibility.}
All results are reported as statistics over eight independent random seeds (42--49). Checkpoints with the best primary validation metric are retained per seed. Ensemble test predictions are produced by averaging sigmoid probabilities (Tasks~2 and~3) or raw scalars (Task~1) across valid seeds. Seeds producing degenerate predictions (test prediction range $< 0.01$) are excluded as collapsed runs.

\subsubsection{Task 3 dual-view baseline.}
The dual-view model processes both views simultaneously by concatenating the preprocessed A4C and A2C clips along the temporal axis prior to feature extraction in the backbone. At inference, test-time augmentation (TTA) with $n=10$ passes of random horizontal flip and temporal jitter is applied, and probabilities are averaged across augmentations and seeds. Only Task~3 labels are used; no Task~1 or Task~2 supervision is introduced.

\subsubsection{Task 3 clinical floor.}
A logistic regression trained on age and baseline LVEF (Model~A-reduced, $N=196$ train+val patients) serves as the clinical reference. This model uses restricted registry variables unavailable during image-based inference and is evaluated strictly to provide a clinical performance floor. It is calibrated with 5-fold cross-validated sigmoid calibration and evaluated with 1,000-iteration bootstrap 95\% CIs.

\subsection{Results}
\label{subsec:results}

\subsubsection{Tasks 1 and 2.}
Table~\ref{tab:results_t1t2} reports Tasks~1 and~2 validation and test results across eight seeds. Task~1 achieves a test MAE of 4.98~pp (7-seed ensemble; one seed excluded due to degenerate test predictions, range~$< 0.01$~pp), within the EchoNet-Dynamic reference range of 4--6~pp~\cite{ouyang2020video}. Validation and test performance are consistent, confirming stable generalisation. Task~2 achieves a test AUC of 0.849 (8-seed ensemble), substantially above the positive rate (16.8\%) and reflecting strong discriminative ability for GLS-defined LV dysfunction from cine video.

\begin{table}[ht]
\centering
\caption{Tasks~1 and~2 reference implementation results. Val mean $\pm$ SD
reported across 8 seeds. Test AUC and MAE are 8-seed ensembles
(Task~1: 7 seeds$^\dagger$; one collapsed seed excluded). $\uparrow$ higher
is better; $\downarrow$ lower is better.}
\label{tab:results_t1t2}
\renewcommand{\arraystretch}{1.2}
\begin{tabular}{lcccc}
\hline
\textbf{Task} & \textbf{Model} & \textbf{Metric} &
\textbf{Val (mean$\pm$SD)} & \textbf{Test (95\% CI)} \\
\hline
Task 1 & R2+1D + LSTM & MAE (pp) $\downarrow$ & 5.02 $\pm$ 0.16 & 4.98$^\dagger$ (4.54--5.46) \\
Task~2 & R2+1D + LSTM & AUC $\uparrow$        & $0.755 \pm 0.025$ & $0.849~(0.793\text{--}0.890)$ \\
\hline
\end{tabular}
\end{table}

$^\dagger$ One seed produced degenerate test predictions (range < 0.01 pp) and was excluded; test MAE is a 7-seed ensemble.

\subsubsection{Task 3.}
\begin{figure}[!htp]
\centering
\includegraphics[width=0.55\textwidth]{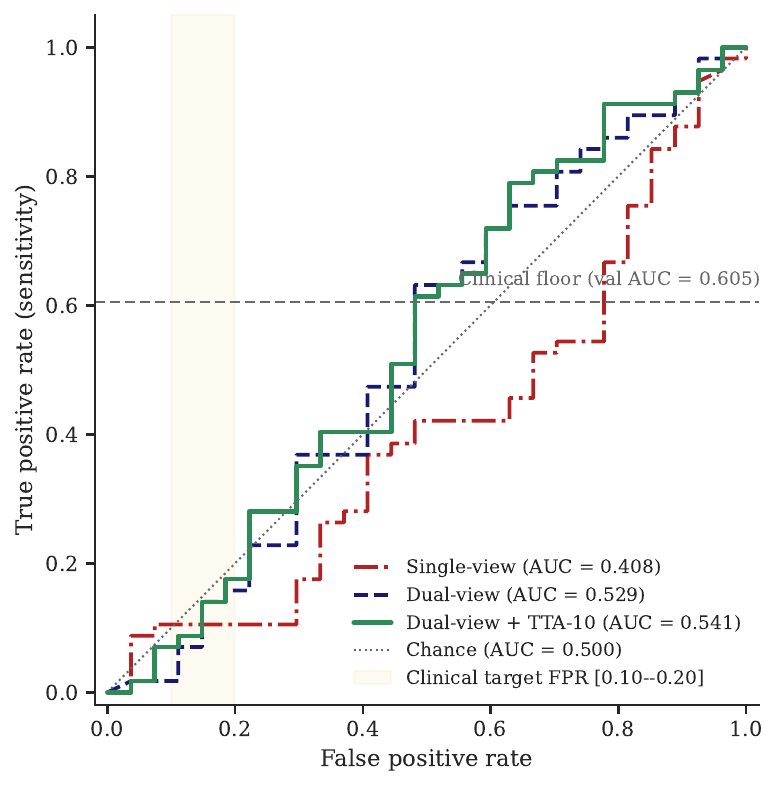}
\caption{Task~3 receiver operating characteristic (ROC) curves evaluated on the test set. The dual-view model enhanced with test-time augmentation (TTA-10) achieves the highest discriminative performance (AUC 0.541). The shaded region indicates the clinical target false positive rate (FPR) of 0.10--0.20, highlighting the operating constraints necessary for viable early risk screening.}
\label{fig:task3_roc}
\end{figure}

Table~\ref{tab:results_t3} and Fig.~\ref{fig:task3_roc} report the full Task~3 benchmark suite. The clinical floor (logistic regression on age and LVEF) achieves val AUC~0.605 (95\%~CI 0.414--0.790) and test AUC~0.525 (95\%~CI 0.388--0.654), confirming that structured clinical variables carry modest but above-chance discriminative signal. The single-view video model achieves val AUC $0.627 \pm 0.049$ but produces near-chance predictions on the test set (AUC 0.408): per-patient predictions collapse to a narrow band, with an output range (max $-$ min) of less than 0.02 probability units and a positive-versus-negative class mean difference of less than 0.002, indicating failure to generalise beyond the validation distribution. The dual-view model, which concatenates A4C and A2C clips along the temporal axis, improves substantially (test AUC~0.529). Adding TTA with $n=10$ augmentation passes further improves to test AUC~0.541.

These results are consistent with internal pilot experiments conducted on an earlier CARDIOCARE data freeze, where the same architecture achieved AUC~0.49--0.54 regardless of pretraining strategy; those results remain unpublished and are reported here solely to confirm that the Task~3 performance ceiling is a stable property of the prediction problem rather than an artefact of the current dataset split.

\begin{table}[!ht]
\centering
\caption{Task~3 reference implementation results. Val AUC reported as mean~$\pm$~SD across 8~seeds (clinical floor: bootstrap 95\%~CI on val). Test AUC is the 8-seed ensemble with bootstrap 95\%~CIs (1,000 iterations); the clinical floor test AUC uses the same bootstrap procedure applied to the held-out test set. $\dagger$~Clinical floor uses restricted registry variables unavailable during image-based inference. \ddag\ Single-view model predictions are near-constant on the test set: output range $<0.02$ probability units, pos--neg mean difference $<0.002$, indicating failure to generalise.}
\label{tab:results_t3}
\renewcommand{\arraystretch}{1.2}
\begin{tabular}{lcc}
\hline
\textbf{Model} & \textbf{Val AUC} & \textbf{Test AUC (95\% CI)} \\
\hline
Clinical floor (age + LVEF)$^\dagger$
  & $0.605~(0.414\text{--}0.790)$
  & $0.525~(0.388\text{--}0.654)$ \\
Video single-view (Kinetics)
  & $0.627 \pm 0.049$
  & $0.408~(0.279\text{--}0.534)^\ddagger$ \\
Video dual-view (Kinetics)
  & $0.646 \pm 0.048$
  & $0.529~(0.379\text{--}0.676)$ \\
Video dual-view + TTA-10
  & $0.646 \pm 0.048$
  & $\mathbf{0.541}~(0.390\text{--}0.692)$ \\
\hline
\end{tabular}
\end{table}

\section{Discussion}
\label{sec:discussion}

Table~\ref{tab:landscape} summarises the current performance landscape across all three tasks. Tasks~1 and~2 validate the dataset integrity and confirm that standard spatiotemporal architectures extract robust functional signal from echocardiography cine loops. Task~3 remains an open problem: the strongest video baseline (AUC~0.541) and the clinical floor (AUC~0.525) achieve statistically indistinguishable performance on the test set, with overlapping 95\% confidence intervals reflecting the inherent difficulty of predicting a future outcome from a single pre-therapy acquisition at the current training scale of $N\!=\!157$.
Given the substantial overlap between the two confidence intervals, this result should be interpreted as evidence that imaging-based and clinical-variable-based approaches achieve comparable, near-chance discriminative performance at the current sample size, rather than as evidence that one approach is reliably superior to the other.

\begin{table}[!ht]
\centering
\caption{EchoRisk performance landscape. Test results are 8-seed
ensembles with bootstrap 95\% CIs. The clinical floor (Task~3) uses
restricted registry variables and is excluded from participant ranking.}
\label{tab:landscape}
\renewcommand{\arraystretch}{1.2}
\begin{tabular}{llcc}
\hline
\textbf{Task} & \textbf{Model} & \textbf{Primary metric} &
\textbf{Test (95\% CI)} \\
\hline
Task~1 & R2+1D + LSTM & MAE $\downarrow$ (pp)
  & $4.98~(4.54\text{--}5.46)$ \\
Task~2 & R2+1D + LSTM             & AUC $\uparrow$
  & $0.849~(0.793\text{--}0.890)$ \\
Task~3 & Video dual-view + TTA-10 & AUC $\uparrow$
  & $0.541~(0.390\text{--}0.692)$ \\
Task~3 & Clinical floor$^\dagger$ & AUC $\uparrow$
  & $0.525~(0.388\text{--}0.654)$ \\
\hline
\end{tabular}
\end{table}

Important open questions for the community that can be explored with the EchoRisk dataset include how to improve representation learning specifically for the domain of cardiac ultrasound, how to better capture complex temporal dynamics and longitudinal changes across multiple patient visits, and the optimal strategies for integrating spatial information from paired apical views. Developing architectures capable of reasoning over variable-length sequences to model subtle functional decline remains a significant avenue for research.

While this dataset provides a comprehensive foundation for predicting therapy-induced cardiac dysfunction from video, early cardiotoxicity remains a multifactorial outcome. Future work could explore the fusion of imaging features with available clinical metadata, such as baseline risk factors and specific treatment regimens, to construct holistic predictive models. Additionally, adapting methodologies to account for variable follow-up horizons and outcome heterogeneity will be crucial for approaching clinical performance thresholds and prognosticating long-term cardiovascular risk in oncology patients.

\FloatBarrier

\bibliographystyle{splncs04}
\bibliography{mybibliography}

\end{document}